\documentclass[]{TEAI}
\usepackage{helvet}

\usepackage{amsmath} 
\usepackage{natbib}
\usepackage{graphicx}
\usepackage{subcaption} 

\usepackage[toc,page,header]{appendix}
\usepackage[utf8]{inputenc} 
\usepackage[T1]{fontenc}    
\usepackage{hyperref}       
\usepackage{url}            
\usepackage{xspace}
\usepackage{booktabs}       
\usepackage{lmodern}        
\usepackage{amsfonts}       
\usepackage{nicefrac}       
\usepackage{microtype}      
\usepackage{wrapfig}

\usepackage{amssymb}  
\usepackage{fontawesome}  
\usepackage{url}  

\usepackage{titletoc}

\usepackage{tikz}  
\usepackage{comment}  
\usepackage{tabularx}  
\usepackage{booktabs}  

\usepackage{minitoc}

\usepackage{booktabs}
\usepackage{array}
\usepackage{etoolbox}

\definecolor{lightblue}{RGB}{200, 230, 255}  
\definecolor{headerblue}{RGB}{150, 200, 255} 

\usepackage{pgfplots}
\usepackage[utf8]{inputenc} 
\usepackage[T1]{fontenc}    
\usepackage{hyperref}       
\usepackage{url}            
\usepackage{booktabs}       
\usepackage{amsfonts}       
\usepackage{nicefrac}       
\usepackage{microtype}      
\usepackage{xcolor}         
\usepackage{graphicx}
\usepackage{float}
\usepackage{comment}
\usepackage{multirow} 
\usepackage{amsmath} 
\usepackage{makecell} 
\usepackage{siunitx}  
\usepackage{tikz}
\usepackage{pgf-pie} 
\usepackage{subcaption}
\usepackage{wrapfig}
\usepackage[export]{adjustbox}

\usepackage{ragged2e}      
\usepackage{tabularx}       
\usepackage{array}          
\usepackage{caption}        
\usepackage{enumitem}
\usepackage{pifont}
\usepackage[hang,flushmargin]{footmisc} 

\usepackage{tcolorbox}

\usepackage{tcolorbox}    
\tcbuselibrary{breakable}  
\tcbuselibrary{skins}      

\usepackage{tabularx}
\usepackage{listings}

\newcommand{\ourmethod}{\textsc{BiDPO}\xspace}
\newcommand{\ourdataset}{\textsc{BiComp}\xspace}
\newcommand{\fakeparagraph}[1]{\noindent\textbf{#1}}

\newcommand{\up}[1]{{\textcolor{green}{~$\uparrow$#1}}}

\newcommand{\ie}{\textit{i.e.\ }}

\title{Compositional Text-to-Image Generation Via Region-aware Bimodal Direct Preference Optimization}

\author{
    Zhuohan Liu\textsuperscript{1,2,*},
    Wujian Peng\textsuperscript{1,2,*},
    Yitong Chen\textsuperscript{1,2},  
    Zuxuan Wu\textsuperscript{1,2,$\dagger$}
}

\affiliation[1]{\mbox{Shanghai Key Lab of Intell. Info. Processing, School of CS, Fudan University}} 
\affiliation[2]{\mbox{Shanghai Collaborative Innovation Center of Intelligent Visual Computing}}

\abstract{
\begin{abstract}

Despite the rapid progress of text-to-image (T2I) models, generating images that accurately reflect complex compositional prompts (covering attribute bindings, object relationships, counting) still remains challenging. To address this, we propose \ourmethod, a framework to enhance T2I model's capability of compositional text-to-image generation. We begin by introducing an carefully designed pipeline to construct a large-scale preference dataset, \ourdataset, with strictly quality control. Then, we extend Diffusion DPO to jointly optimize image and text preferences, which is shown to greatly effective in improving the models to follow complex text prompt in generation. To further enhance the models for fine-grained alignment, we employ a region-level guidance method to focus on regions relevant to compositional concepts. Experimental results demonstrate that our \ourmethod substantially improves compositional fidelity, consistently outperforming prior methods across multiple benchmarks. Our approach highlights the potential of preference-based fine-tuning for complex text-to-image tasks, offering a flexible and scalable alternative to existing techniques.
Code is available at \url{https://github.com/anzeameol/BiDPO}.
\end{abstract}
}

\begin{document}
\maketitle
\renewcommand{\thefootnote}{}
\footnotetext{$^*$Equal Contribution.\\$^\dagger$Corresponding authors.}
\renewcommand{\thefootnote}{\arabic{footnote}}

\vspace{-1.5em}

\section{Introduction}
\label{sec:intro}

\begin{figure}[htbp]
    \centering
    \includegraphics[width=0.8\textwidth]{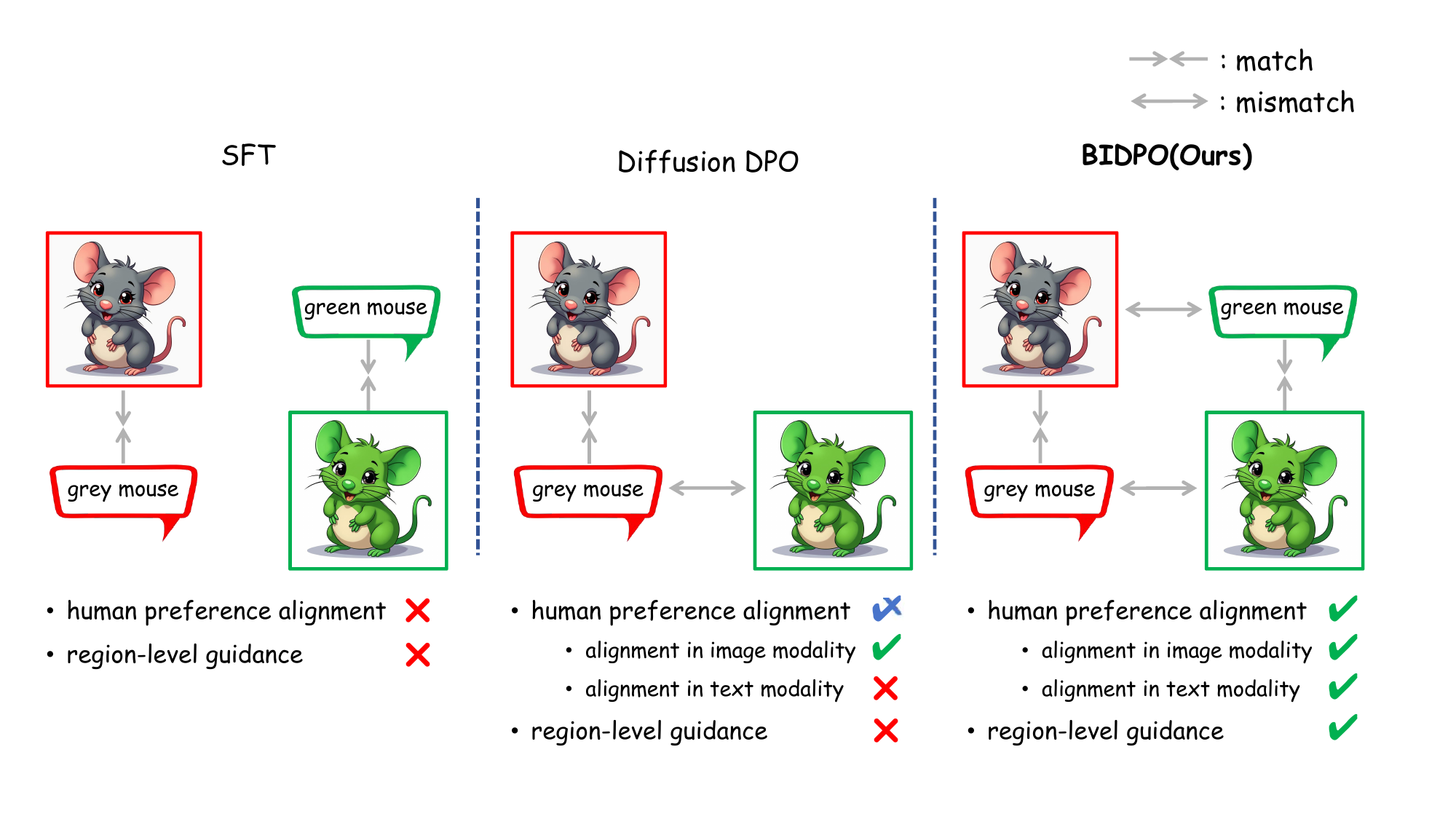}
    \caption{\textbf{Comparison of post-training optimization methods used in compositional text-to-image generation.} Our proposed \ourmethod, achieves full human preference alignment across both image and text modalities while offering region-level guidance, outperforming existing approaches such as SFT, DiffusionDPO.}
    \label{fig:teaser}
\end{figure}

Text-to-Image~(T2I) generation has witnessed remarkable advancements in recent years, largely driven by the rapid development of diffusion models~\citep{dit,sd3,dalle3,flux}. 
While existing models excel at generating images with high fidelity and aesthetics quality, they still struggle to accurately follow complex text instructions, especially when there are multiple objects, different attributes binding to each object, and complex inter-object relationships like spatial relationships~\citep{huang2023t2icompbench}.

To address these challenges, the research community has explored a variety of strategies. Some previous works introduce additional modalities, such as layouts~\citep{Zhang2024CreatiLayoutSM, Patel2024EnhancingIL, Xie2023BoxDiffTS, Chefer2023AttendandExciteAS, Wang2023TokenComposeTD}, scene graphs~\citep{Li2024LAIONSGAE}, or semantic panels~\citep{Feng2023RanniTT} to provide structural guidance for the image generation process. While these approaches have achieved notable improvements, they heavily relies on supplementary inputs that may be difficult to obtain in practice. Another line of work seeks to enhance model comprehension through the integration of Large Language Models~\citep{Lian2023LLMgroundedDE} as a tool; however, such methods can be unstable and computationally intensive. 
\textbf{Motivated by this, we aim to enhance the compositional generation ability under pure text conditions}, without relying on external tools or modalities.

Direct Preference Optimization (DPO)~\citep{Rafailov2023DirectPO}, a powerful variant of Reinforcement Learning from Human Feedback (RLHF), refines traditional reward-model-based RLHF methods and has shown considerable promise in aligning generative models with human preferences. Despite its potential, the application of DPO to compositional text-to-image generation remains largely unexplored. We posit that DPO is particularly well-suited for this domain, as it can effectively leverage human feedback to enhance a model’s ability to interpret and generate intricate compositions. Importantly, as a post-training technique, DPO can be applied to any pre-trained text-to-image model without requiring additional inputs or substantial architectural modifications, thereby offering a simple yet flexible and efficient solution.

In this work, we introduce \ourmethod, a novel framework that employs Bimodal Direct Preference Optimization to advance compositional text-to-image generation. Our approach is distinguished by a fully automated data pipeline for generating high-quality preference data, comprising the following stages: (1) collecting composition-related captions from diverse sources and generating corresponding images using a pre-trained text-to-image model; (2) regenerating captions for these images via a pipeline that integrates object detection, segmentation, and labeling; (3) editing the regenerated captions to produce distinct variants and utilizing an image editing model to modify the original images accordingly; and (4) applying a VQA-based filtering step to ensure the fidelity of the resulting image-caption pairs. The resulting dataset is characterized by high quality, diversity, large scale, and minimal visual differences between preference pairs—attributes essential for effective DPO training.

Subsequently, we extend Diffusion DPO~\citep{Wallace2023DiffusionMA} to a bimodal formulation that jointly considers image and text preferences, and employ this method to fine-tune a pre-trained Stable Diffusion model on the generated preference data. To further enhance model robustness and realism, we incorporate real-world data from VisMin~\citep{Awal2024VisMinVM} dataset, thereby increasing the diversity and authenticity of the training corpus. Additionally, we introduce a region-aware training loss that accentuates specific regions of the image corresponding to captions. This, in conjunction with minimal visual differences in other regions, enables the model to more effectively learn and apply compositional modifications. Experimental results on T2I-CompBench~\citep{huang2023t2icompbench} shows that our method leads to an average of 17\% improvement in ``attribute binding'' category and a overall 10\% improvement over the base model, demonstrating the effectiveness of our approach.

Our contributions are summarized as follows:
\begin{itemize}[leftmargin=1em]
    \item We first introduce DPO to compositional text-to-image generation by presenting \ourmethod, a novel framework that improves model alignment through fine-grained preference optimization on both text and image modalities.
    
    \item We propose a region-level guidance mechanism that selectively steers the model's focus toward regions of interest. This mechanism is shown to substantially enhance the capability for fine-grained text-to-image alignment.
    
    \item We developed an automated data pipeline to construct a large-scale, high-quality text-to-image preference dataset, which includes both textual and visual negative examples. The proposed \ourdataset comprise 57,474 original images and 94,502 edited images, covering six dimensions: color, shape, texture, spatial relationship, non-spatial relationship and numeracy.
    
    \item We conducted extensive experiments on several widely-used benchmarks, demonstrating significant performance gains over previous state-of-the-art methods.
    
\end{itemize}

\section{Related Works}

\subsection{Compositional Text-to-Image Generation}
The field of text-to-image (T2I) generation has undergone rapid progress with the emergence of large-scale diffusion models. These models are capable of synthesizing highly realistic images conditioned on textual prompts, and recent systems such as Stable Diffusion 3~\citep{sd3}, DALL-E 3~\citep{dalle3}, and Flux~\citep{flux} have achieved strong performance on standard quality benchmarks. Nevertheless, accurately capturing compositional semantics---involving multiple objects, attributes, and relations---remains a persistent challenge. Recent benchmark studies, including T2I-CompBench~\citep{huang2023t2icompbench}, GenEval~\citep{ghosh2023geneval} and DPG-Bench~\citep{dpgbench}, highlight that state-of-the-art models often fail on fine-grained object binding and spatial reasoning tasks. Multiple methods have been proposed to address these limitations, such as incorporating structured scene representations ~\citep{Feng2023RanniTT,Zhang2024CreatiLayoutSM,Li2024LAIONSGAE}, conducting more precise control by generating the foreground objects and background separately \citep{Xie2023BoxDiffTS,Lian2023LLMgroundedDE}, leveraging large vision-language models or multimodal LLM to improve understanding \citep{Lian2023LLMgroundedDE, Yang2024MasteringTD}, guiding the image-text cross-attention activations \citep{Chefer2023AttendandExciteAS}, employing contrastive learning techniques \citep{contrafusion}, and introducing reinforcement learning strategies \citep{zhang2024itercomp}. Our work complements these approaches by focusing on preference-based optimization techniques to further align T2I models with human expectations on compositional tasks.

\subsection{Preference Alignment in Image Synthesis}

Preference alignment has become a central strategy for bridging the gap between model generations and human expectations. Early approaches adapt Reinforcement Learning from Human Feedback (RLHF), which is originally developed for large language models, to the image domain by training reward models or synthetic comparisons and optimizing with on-policy algorithms such as PPO \citep{Lee2023AligningTM, Xu2023ImageRewardLA}. However, RLHF pipelines are computationally expensive and unstable when applied to high-dimensional image spaces. To address these limitations, DPO \citep{Rafailov2023DirectPO} was proposed as a simpler, more stable alternative that bypasses reinforcement learning by directly optimizing a contrastive preference objective. While DPO was first studied in language generation, recent works have begun adapting it to diffusion models, showing promising improvements in human preference alignment\citep{Wallace2023DiffusionMA, Lee2025CalibratedMO, Karthik2024ScalableRP, Hong2024MarginawarePO, Liang2024AestheticPD, Zhang2025DiffusionMA, Zhu2025DSPODS}. These results suggest that preference-based optimization without explicit reward modeling provides a practical pathway for fine-grained alignment in image synthesis. However, existing studies primarily focus on overall image quality and safety, with limited exploration of compositional capabilities. Our work extends the application of DPO to compositional T2I tasks, demonstrating that it can effectively enhance models' abilities to handle complex object interactions and attributes.

\section{Method}
\subsection{Diffusion DPO.}

Diffusion DPO~\citep{Wallace2023DiffusionMA} is a recent advancement in the
field of diffusion models, which applies the principles of DPO to enhance the
training of diffusion models. The core idea is to leverage human feedback in
the form of preference data to guide the model towards generating outputs that
are more aligned with human preferences. In Diffusion DPO, the training loss is
defined as:
\begin{align}
    \mathcal{L}\bigl(\theta\bigr) =
     & -\mathbb{E}_{\bigl({\boldsymbol{x}}_0^w, {\boldsymbol{x}}_0^l\bigr) \sim \mathcal{D},\ t \sim \mathcal{U}\bigl(0,T\bigr),\ {\boldsymbol{x}}_t^w \sim q\bigl({\boldsymbol{x}}_t^w|{\boldsymbol{x}}_0^w\bigr),\ {\boldsymbol{x}}_t^l \sim q\bigl({\boldsymbol{x}}_t^l|{\boldsymbol{x}}_0^l\bigr)} \nonumber \\
     & \log \sigma\bigl( -\beta T \omega\bigl(\lambda_t\bigr) \bigr) \bigl( \nonumber \\
     & \quad \| \boldsymbol{\epsilon}^w - \boldsymbol{\epsilon}_\theta\bigl({\boldsymbol{x}}_t^w, t\bigr) \|_2^2 - \| \boldsymbol{\epsilon}^w - \boldsymbol{\epsilon}_{\mathrm{ref}}\bigl({\boldsymbol{x}}_t^w, t\bigr) \|_2^2 \nonumber \\
     & \quad - \bigl( \| \boldsymbol{\epsilon}^l - \boldsymbol{\epsilon}_\theta\bigl({\boldsymbol{x}}_t^l, t\bigr) \|_2^2 - \| \boldsymbol{\epsilon}^l - \boldsymbol{\epsilon}_{\mathrm{ref}}\bigl({\boldsymbol{x}}_t^l, t\bigr) \|_2^2 \bigr) \bigr)
\end{align}
where \( \mathcal{D} \) is the dataset of preference pairs, \(
{\boldsymbol{x}}_0^w \) and \( {\boldsymbol{x}}_0^l \) are the preferred and
less preferred samples respectively, \( t \) is a randomly sampled time step,
\( q\left({\boldsymbol{x}}_t|{\boldsymbol{x}}_0\right) \) is the forward diffusion
process, \( \boldsymbol{\epsilon}_\theta \) is the model's noise prediction, \(
\boldsymbol{\epsilon}_{\mathrm{ref}} \) is the reference model's noise
prediction, \( \beta \) is a scaling factor, and \( \omega\left(\lambda_t\right) \) is a
weighting function based on the noise level at time step \( t \).

\subsection{\ourmethod}

\begin{figure}[!t]
    \centering
    \begin{subfigure}{0.8\textwidth}
        \centering
        \includegraphics[width=0.8\textwidth]{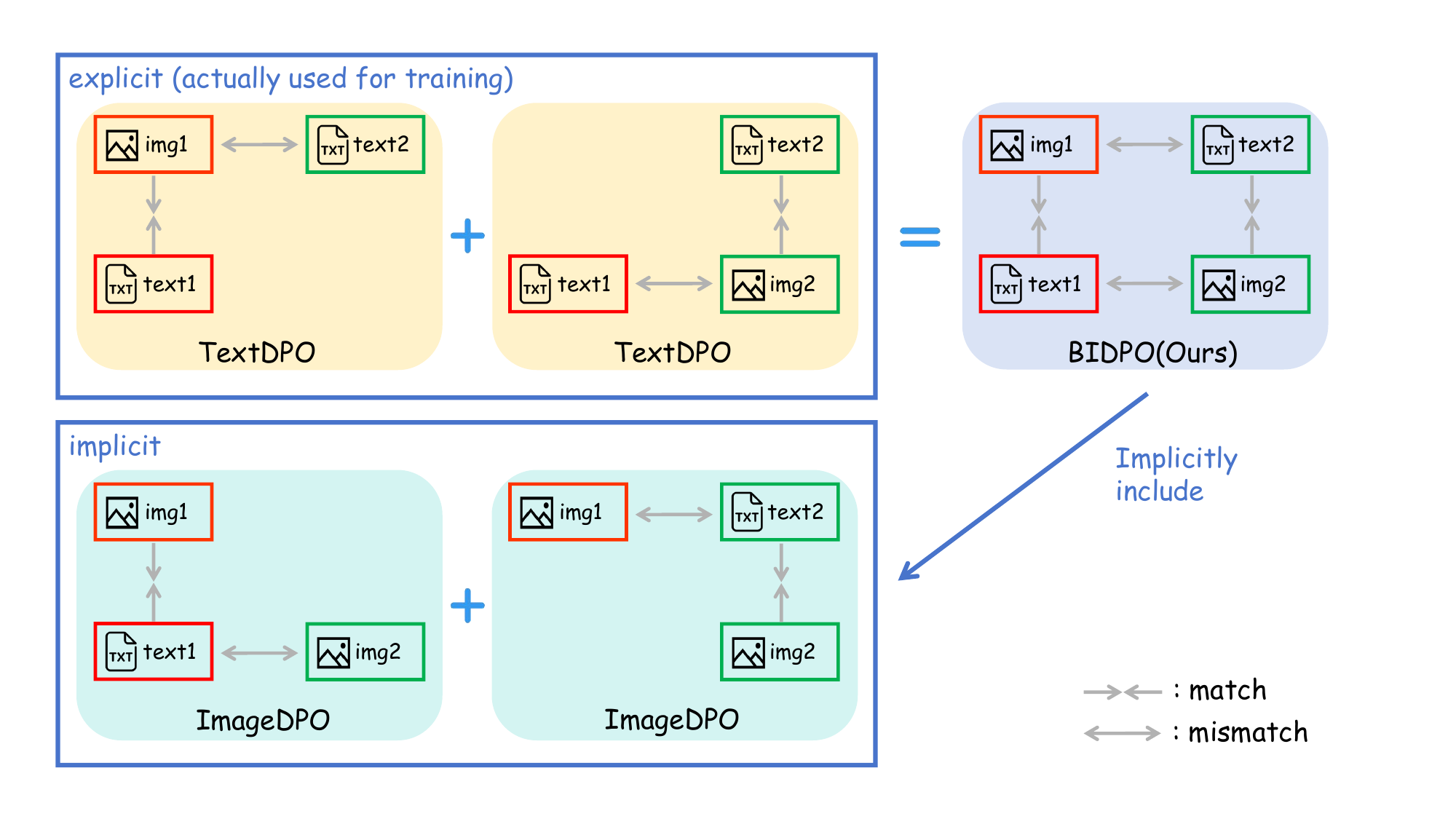}
        \caption{This picture vividly shows how the preference alignment in image modality is implicitly achieved through two explicit text modality preference alignments.}
        \label{fig:method1}
    \end{subfigure}

    \begin{subfigure}{\textwidth}
        \centering
        \includegraphics[width=0.8\textwidth]{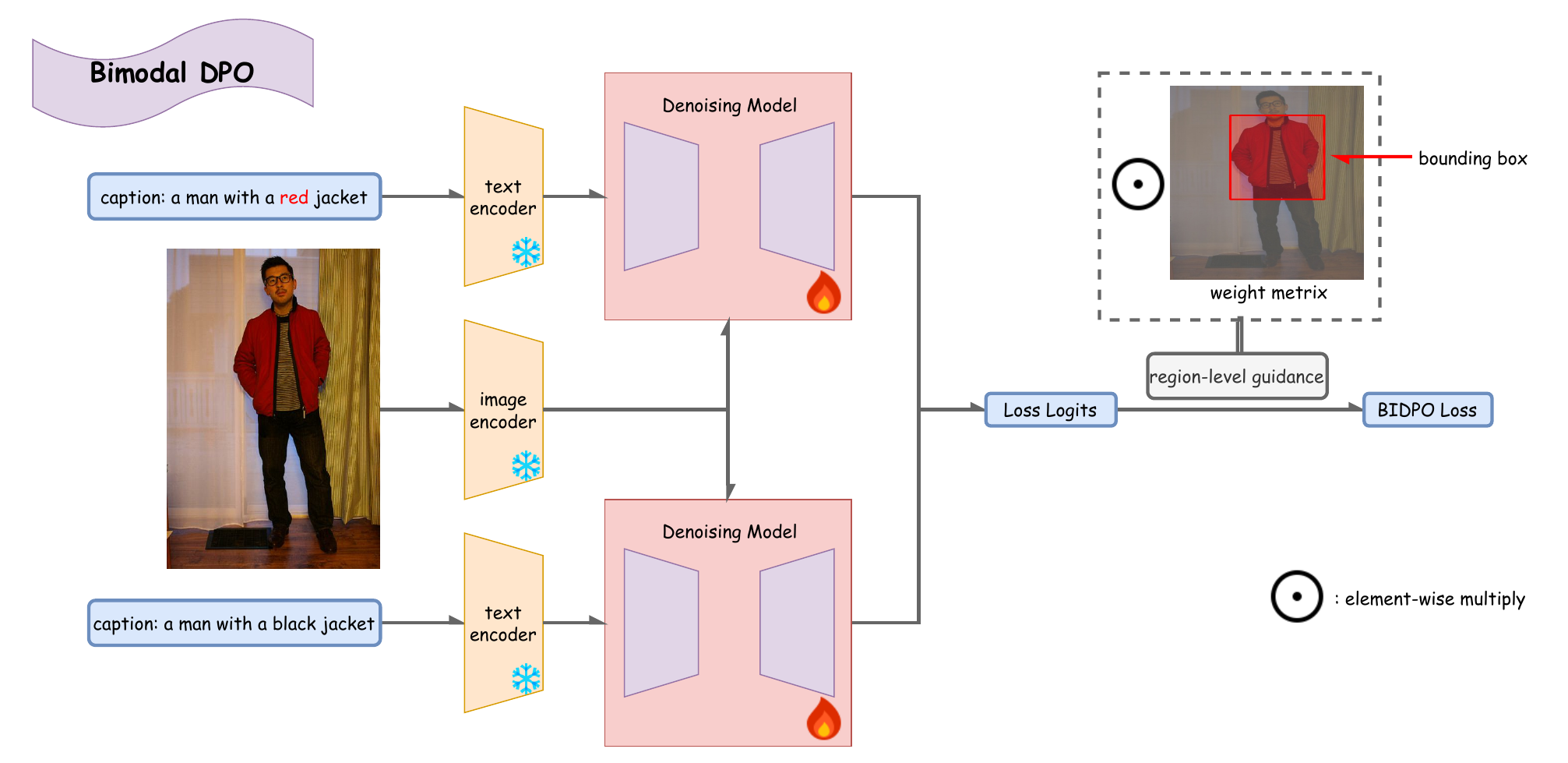}
        \caption{This picture shows one \textbf{TextDPO} process with region-level guidance. \textbf{TextDPO}, as an extension of Diffusion DPO, keep the prefered sample to be the preferred image with the preferred caption, while altering the less prefered sample to be the preferred image with the less preferred caption. The region-level guidance is applied during the loss calculation to guide the model to focus on the most relevant regions.}
        \label{fig:method2}
    \end{subfigure}
    
    \caption{\textbf{Overview of our proposed \ourmethod.} (a) \ourmethod integrates bimodal(image and text) preference alignment; (b) Diffusion process and loss calculation with region-level guidance.}
    \label{fig:bidpo}
\end{figure}

\fakeparagraph{Bimodal DPO.}
While Diffusion DPO \citep{Wallace2023DiffusionMA} have demonstrated promising results in
preference learning, it suffers from several critical limitations in handling
compositional and complex visual scenes. \textbf{First}, current methods
primarily focus on image-to-image contrastive learning while largely ignoring
the textual modality. This represents a significant drawback given that textual
understanding plays a crucial role in compositional reasoning \citep{contrafusion}.
\textbf{Second}, existing approaches lack regional guidance mechanisms for
complex scenes. When presented with intricate visual compositions, these
methods perform global contrastive learning without explicitly directing
attention to the specific regions that require comparative analysis. To address
these limitations, we extend Diffusion DPO and propose \ourmethod, which
integrates bimodal contrastive learning and region-level guidance.

We first extend the Diffusion DPO to a text-based version that focuses on text
preferences, denoted as \textbf{TextDPO}. Based on the idea that Diffusion DPO basically depresses the
diffusion process of the less preferred sample while enhancing the diffusion
process of the preferred sample, we alter the less prefered sample to be the
preferred image with the less preferred caption. The training loss is defined
as:
\begin{equation}
\begin{gathered}[b]
\mathcal{L}_{\text{TextDPO}}\bigl(\theta\bigr) =
 -\mathbb{E}_{\bigl({\boldsymbol{x}}_0^w, {\boldsymbol{y}}^w, {\boldsymbol{y}}^l\bigr) \sim \mathcal{D},\ t \sim \mathcal{U}\bigl(0,T\bigr),\ {\boldsymbol{x}}_t^w \sim q\bigl({\boldsymbol{x}}_t^w|{\boldsymbol{x}}_0^w\bigr)} \\
 \log \sigma\bigl( -\beta T \omega\bigl(\lambda_t\bigr) \bigr) \bigl( \vphantom{-} \\
 \| \boldsymbol{\epsilon}^w - \boldsymbol{\epsilon}_\theta\bigl({\boldsymbol{x}}_t^w, t, c^w\bigr) \|_2^2 - \| \boldsymbol{\epsilon}^w - \boldsymbol{\epsilon}_{\mathrm{ref}}\bigl({\boldsymbol{x}}_t^w, t, c^w\bigr) \|_2^2 \\
 - \bigl( \| \boldsymbol{\epsilon}^l - \boldsymbol{\epsilon}_\theta\bigl({\boldsymbol{x}}_t^w, t, c^l\bigr) \|_2^2 - \| \boldsymbol{\epsilon}^l - \boldsymbol{\epsilon}_{\mathrm{ref}}\bigl({\boldsymbol{x}}_t^w, t, c^l\bigr) \|_2^2 \bigr) \bigr)
\end{gathered}
\end{equation}
where \( \mathcal{D} \) is the dataset of preference pairs, \( {\boldsymbol{x}}_0^w \) is the preferred image, \( {\boldsymbol{y}}^w \) and \( {\boldsymbol{y}}^l \) are the preferred and less preferred captions respectively, \( t \) is a randomly sampled time step, \( q\left({\boldsymbol{x}}_t|{\boldsymbol{x}}_0\right) \) is the forward diffusion process, \( \boldsymbol{\epsilon}_\theta \) is the model's noise prediction which also conditioned on text embeddings \( c^w \) and \( c^l \), \( \boldsymbol{\epsilon}_{\mathrm{ref}} \) is the reference model's noise prediction, \( \beta \) is a scaling factor, and \( \omega\left(\lambda_t\right) \) is a weighting function based on the noise level at time step \( t \).

We then construct \ourmethod by combining \textbf{two TextDPO procedures}.
For two image-caption pairs with slight difference
\(\left({\boldsymbol{x}}_0^w, {\boldsymbol{y}}^w\right)\) and \(\left({\boldsymbol{x}}_0^l,
{\boldsymbol{y}}^l\right)\), we create two training samples, \(\left({\boldsymbol{x}}_0^w, {\boldsymbol{y}}^w, {\boldsymbol{y}}^l\right)\) and \(\left({\boldsymbol{x}}_0^l, {\boldsymbol{y}}^l, {\boldsymbol{y}}^w\right)\), each of which is used to compute a TextDPO loss. 
This way, through the TextDPO loss, the model learns to prefer caption \(
{\boldsymbol{y}}^w \) over \( {\boldsymbol{y}}^l \) for image \(
{\boldsymbol{x}}_0^w \), which means that image \( {\boldsymbol{x}}_0^w \) and
caption \( {\boldsymbol{y}}^l \) are the less preferred pair and this diffusion
process should be depressed. Similarly, through the second training sample, the
model learns to prefer caption \( {\boldsymbol{y}}^l \) over \(
{\boldsymbol{y}}^w \) for image \( {\boldsymbol{x}}_0^l \), which means that
image \( {\boldsymbol{x}}_0^l \) and caption \( {\boldsymbol{y}}^l \) are the
preferred pair and this diffusion process should be enhanced.
By combining these two losses, we effectively establish that image \( {\boldsymbol{x}}_0^l \) and caption \( {\boldsymbol{y}}^l \) form the preferred pair, while image \( {\boldsymbol{x}}_0^w \) and caption \( {\boldsymbol{y}}^l \) constitute the less preferred pair,
which guides the model to learn to
prefer image \( {\boldsymbol{x}}_0^l \) over \( {\boldsymbol{x}}_0^w \) for
caption \( {\boldsymbol{y}}^l \). And also the same way, the model learns to prefer image \( {\boldsymbol{x}}_0^w \) over \( {\boldsymbol{x}}_0^l
\) for caption \( {\boldsymbol{y}}^w \). Therefore, the image-to-image contrastive learning is implicitly achieved through the combination of two explicit text-to-text contrastive learning processes, as shown in \Cref{fig:method1}.

\fakeparagraph{Region-level Guidance for Fine-grained Alignment.}
To further enhance the model's ability to focus on specific regions of the image that correspond to the edited captions, we introduce a region-level guidance method. This method adjusts the importance of different regions in the image during training, helping the model to better understand and learn the desired modifications.
We define the region-level guidance method as follows:
\begin{equation}
    \mathcal{L}_{\text{BIDPO-region}}\left(\theta\right) = \mathcal{L}_{\text{BIDPO}}\left(\theta\right) \odot M
\end{equation}
where \( M \) is a mask that highlights the regions of the image corresponding to the edited captions, and the operator \( \odot \) denotes element-wise multiplication.
The mask is generated according to the bounding boxes of the objects involved in the edits, which are obtained from the caption generation and editing pipeline. We set a smaller weight for the regions not involved in the edits, ensuring that the loss is focused on the relevant regions of the image.

\begin{figure}[!t]
    \centering
    \includegraphics[width=1.0\textwidth]{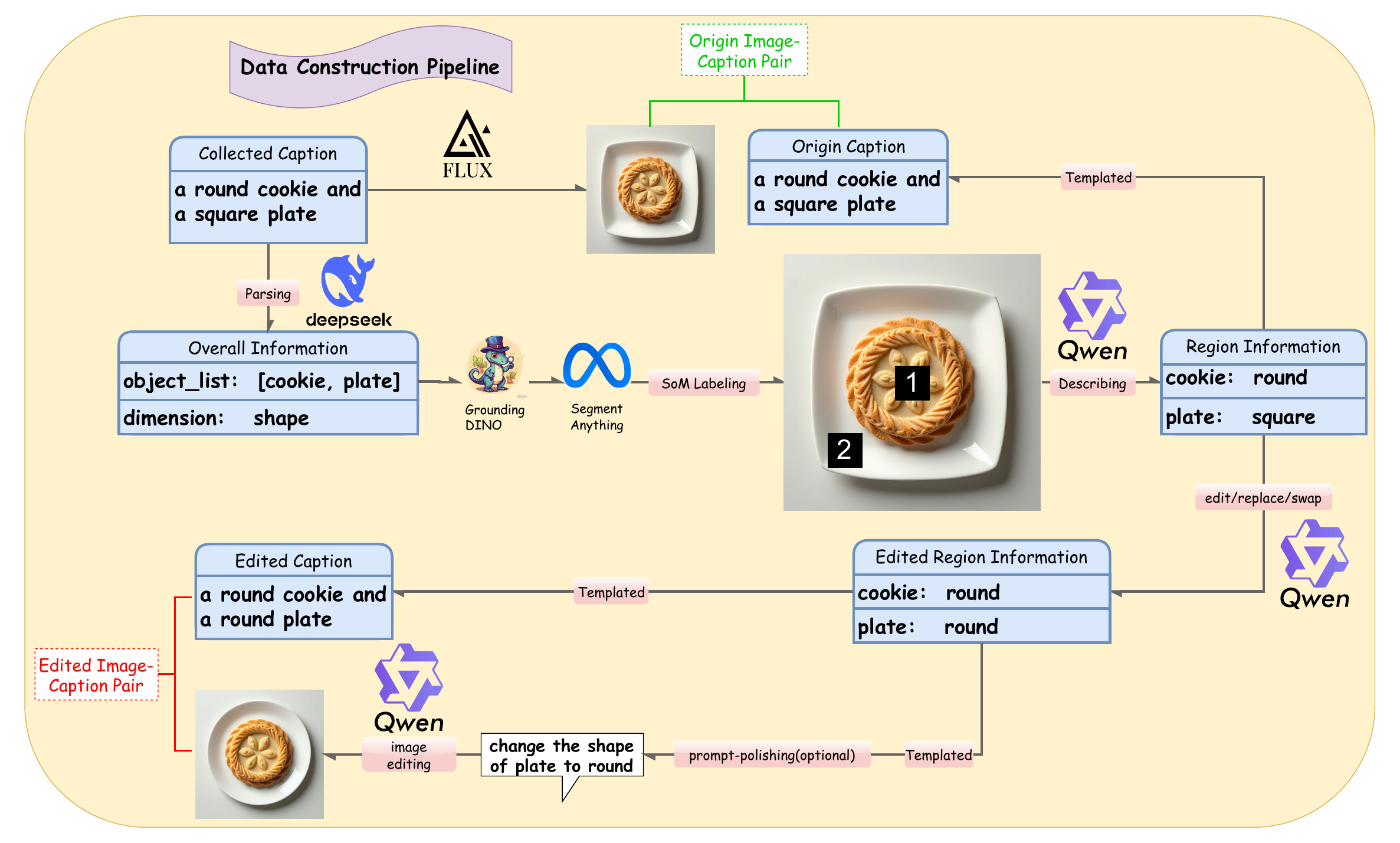}
    \caption{\textbf{The data construction pipeline of our \ourdataset dataset.} Our \ourdataset dataset, though generated automatically, contains large amounts of high-quality image-caption pairs with region annotations across multiple composition-related dimensions.}
    \label{fig:data_pipeline}
\end{figure}

\subsection{Data Pipeline}

Given the absence of publicly available, high-quality region-annotated composition preference datasets suitable for \ourmethod training, we design a data generation pipeline to construct \ourdataset, a large-scale, high-quality dataset with regional annotations.

\fakeparagraph{Prompt Collection and Image Generation.}
We collect composition-related captions from various sources, including: CONPAIR~\citep{contrafusion}, ReasonGen-R1~\citep{Zhang2025ReasonGenR1CF}, T2I-R1~\citep{Jiang2025T2IR1RI}, T2I-CompBench Test Set~\citep{huang2023t2icompbench}. For each collected caption, we generate 2-4 images using Flux.1-dev~\citep{flux}.

\fakeparagraph{Caption Generation.}
Considering that the generated images may not always perfectly align with the original captions, we employ a caption generation pipeline to create new captions that better describe the generated images. The pipeline includes the following steps:
\begin{itemize}[leftmargin=1em]
    \item \textbf{Dimension Parsing:} We use DeepSeek-V3~\citep{DeepSeekAI2024DeepSeekV3TR} to parse the original captions and identify which dimension the caption is referring to (``color'', ``shape'', ``texture'', ``spatial'', ``action'', ``numeracy'' or ``other''). If the caption refer to multiple dimensions, we select one with the following priority (from highest to lowest): object relationship(spatial, action), numeracy, attribute binding(color, shape, texture). If the caption does not refer to any of the specified dimensions, we classify it as ``other''.
    \item \textbf{Object List Parsing:} We use DeepSeek-R1~\citep{DeepSeekAI2025DeepSeekR1IR} to extract the list of objects mentioned in the original captions.
    \item \textbf{Grounding Dino Detection and SAM Segmentation:} We use Grounding Dino~\citep{Liu2023GroundingDM} to detect objects in the generated images based on the object list extracted in the previous step. We then use SAM2~\citep{Ravi2024SAM2S} to segment the detected objects and obtain their masks.
    \item \textbf{VLM Describing:} We use Qwen2.5-VL-72B-Instruct~\citep{Bai2025Qwen25VLTR} to label each segmented object in the image. First we label the image with SoM (Set-of-Mark) masks, which are highlighted regions in the image. Then, we ask Qwen to describe each masked object in detail, including its attributes (e.g., color, shape, texture) or relationships with other objects. We use specific prompts to guide the model based on the dimension identified in the ``Dimension Parsing'' step.
    \item \textbf{Caption Synthesis:} Finally, we synthesize a new caption by combining the labels generated in the previous step. We use a template-based approach to ensure that the new caption is coherent and accurately describes the content of the image. In addition, for the ``numeracy'' dimension, we skip the VLM labeling step and directly use the result from Grounding Dino to count the number of objects and generate a caption accordingly.
\end{itemize}

We also filter out image-caption pairs that contain too many objects, with the
consideration of the bad performance of detection and segmentation models in
such cases and the convenience of the following image editing step.

\begin{table}[t]
    \centering
    \caption{\textbf{Number of images in each dimension.} Each original image may correspond to multiple edited images.}
    \resizebox{1.0\linewidth}{!}{
        \begin{tabular}{l c c c c c c c}
            \toprule
                                    & Color & Shape & Texture & Spatial & Non-spatial & Numeracy & Total \\
            \midrule
            Original Image & 19714 & 5399  & 9728    & 7919    & 3647        & 11067    & 57474 \\
            Edited Image  & 46006 & 8473  & 17345   & 7919    & 3647        & 11112    & 94502 \\
            \bottomrule
        \end{tabular}
    }
    \label{tab:image_counts}
\end{table}

\fakeparagraph{Caption Editing and Image Editing.}
To generate the preference data, we first edit the regenerated captions to create distinct versions. We use  Qwen2.5-VL-72B-Instruct~\citep{Bai2025Qwen25VLTR} to generate distinct region information (attributes, relationships) based on the image with SoMs and the original region information.
Then, we use Qwen-Image-Edit~\citep{Wu2025QwenImageTR} model to edit the original image based on specific prompts. These prompts are designed to reflect the changes made in the edited captions, which are also generated in a template-based manner. For ``action'' and ``numeracy'' dimensions, Considering the complexity of editing images with multiple objects, we enhance the prompts by adding more detailed instructions using Qwen2.5-VL-72B-Instruct.

In order to enhance the model's ability to correctly attribute properties to
objects, we add three more edited captions for each image-caption pair in ``color'', ``shape'', and ``texture'' dimensions when there are exactly \textbf{two} objects:
\begin{itemize}[leftmargin=1em]
    \item Swap the attributes of the two objects. For example, if the original caption is
          ``A red ball and a blue cube'', the edited caption would be ``A blue ball and a
          red cube''.
    \item Replace the attribute of one object with the attribute of the other object.
          For example, if the original caption is ``A red ball and a blue cube'', the
          edited captions would be ``A red ball and a red cube" and "A blue ball and a
          blue cube''.
\end{itemize}

\fakeparagraph{Creatilayout Generation.}
For the ``spatial'' dimension, it is hard to edit the image to reflect the changes in the edited caption. We use a different pipeline to generate the source and edited image-caption pairs.
First, we use DeepSeek-V3~\citep{DeepSeekAI2024DeepSeekV3TR} to parse the original caption and generate a layout that describes the whole scene.
Then, we use DeepSeek-V3~\citep{DeepSeekAI2025DeepSeekR1IR} again to edit the layout to a distinct version which differs in spatial relationships.
Finally, we use CreatiLayout~\citep{Zhang2024CreatiLayoutSM} to generate images based on these layouts.

\fakeparagraph{VQA-based Filtering.}
We employ a VQA-based filtering step to ensure the quality of the generated image-caption pairs. We use Qwen2.5-VL-72B-Instruct~\citep{Bai2025Qwen25VLTR} to answer specific questions about the content of the images based on their captions. If the model's answers do not align with the expected responses, we discard those image-caption pairs. This step helps to ensure that the captions accurately describe the content of the images and that any edits made are reflected in both the images and their corresponding captions.
The final dataset composition is shown in \Cref{tab:image_counts}.
\section{Experiments}

\begin{table}[t]
\centering
\caption{\textbf{Main Results on T2I-CompBench~\citep{huang2023t2icompbench}}.}
\resizebox{0.85\linewidth}{!}{
\begin{tabular}{l ccc cc}
\toprule
\multirow{2}{*}{\textbf{Model}} & 
\multicolumn{3}{c}{\textbf{Attribute Binding}} &
\multicolumn{2}{c}{\textbf{Object Relationship}} \\
\cmidrule(lr){2-4} \cmidrule(lr){5-6}   
& Color & Shape & Texture & Spatial & Non-Spatial \\
\midrule
Stable Diffusion 2~\citep{Rombach2021HighResolutionIS} & 50.65 & 42.21 & 49.22 & 13.42 & 30.96 \\
GLIGEN~\citep{li2023gligen} & 42.88  & 39.98 & 39.04 &  26.32 & 30.36  \\
LMD+~\citep{Lian2023LLMgroundedDE} & 48.14 & 48.65 & 56.99 &  25.37 & 28.28 \\
InstanceDiffusion~\citep{wang2024instancediffusion} & 54.33 & 44.72 & 52.93 & 27.91 & 29.47 \\
Attn-Exct v2~\citep{Chefer2023AttendandExciteAS} & 64.00 & 45.17 &  59.63 & 14.55 & 31.09 \\
PixArt-$\alpha$~\citep{Chen2023PixArtFT} & 68.86 & 55.82 & 70.44 & 20.82 & 31.79 \\
ECLIPSE~\citep{Patel2023ECLIPSEAR} & 61.19 & 54.29 & 61.65 & 19.03 & 31.39 \\
Dimba-G~\citep{Fei2024DimbaTD} & 69.21 & 57.07 & 68.21 & 21.05 & 32.98 \\
GenTron~\citep{Chen2023GenTronDT} & 76.74 & 57.00 & 71.50 & 20.98 & 32.02 \\
GORS~\citep{huang2023t2icompbench} & 66.03 & 47.85 & 62.87 & 18.15 & 31.93 \\
ELLA~\citep{dpgbench} & 72.60 & 56.34 & 66.86 & 22.14 & 30.69 \\
MARS~\citep{He2024MARSMO} & 69.13 & 54.31 & 71.23 & 19.24 & 32.10 \\
EVOGEN~\citep{contrafusion} & 71.04 & 54.57 & 72.34 & 21.76 & 33.08 \\
Flux.1-dev~\citep{flux} & 76.35 & 51.10 & 62.79 & 28.02 & 30.80 \\
\midrule
SDXL~(baseline) & 58.90 & 46.90 & 53.13 & 21.23 & 31.20  \\
\rowcolor{blue!5}
SDXL-\ourmethod & 79.35\up{20.4} & 60.47\up{13.6} & 71.36\up{18.2} & 23.41\up{2.2} & 32.29\up{1.1} \\
\bottomrule
\end{tabular}
}
\label{tab:t2i_compbench_result}
\end{table}

\begin{table}[t]
    \centering
    \caption{\textbf{Main Results on GenEval~\citep{ghosh2023geneval}}.}
    \resizebox{\linewidth}{!}{
    \begin{tabular}{lccccccc}
        \toprule
        \multicolumn{1}{c}{\textbf{Model}} & \multicolumn{1}{c}{Single Obj.} & Two Obj. & Counting & Colors & Position & Color Attri. & \textbf{Overall} \\
        \midrule
        SDv2.1~\citep{Rombach2021HighResolutionIS} & 0.98 & 0.51 & 0.44 & 0.85 & 0.07 & 0.17 & 0.50 \\
        PlayGroundv2.5~\citep{Li2024PlaygroundVT} & 0.98 & 0.77 & 0.52 & 0.84 & 0.11 & 0.17 & 0.56 \\
        Show-o~\citep{Xie2024ShowoOS} & 0.95 & 0.52 & 0.49 & 0.82 & 0.11 & 0.28 & 0.53 \\
        Emu3-Gen~\citep{Wang2024Emu3NP} & 0.98	& 0.71 & 0.34 & 0.81 & 0.17	& 0.21 & 0.54 \\
        FLUX~\citep{flux} & 0.98 & 0.81 & 0.74 & 0.79 & 0.22 & 0.45 & 0.66 \\
        DALL-E 3~\citep{dalle3} & 0.96	& 0.87 & 0.47 & 0.83 & 0.43	& 0.45 & 0.67 \\
        \midrule
        SDXL (baseline) &   0.95 & 0.68 & 0.42 & 0.85 & 0.11 & 0.19 & 0.53 \\
        \rowcolor{blue!5}
        SDXL-\ourmethod    & 1.00\up{0.05}  & 0.86\up{0.18}  & 0.59\up{0.17}  & 0.88\up{0.03}  & 0.19\up{0.08}  & 0.22\up{0.03}  & 0.62\up{0.09} \\
        \bottomrule
    \end{tabular}
    }
    \label{tab:geneval_result}
\end{table}

\begin{table}[t]
\centering
\caption{\textbf{Main Results on DPG-Bench~\citep{dpgbench}}.}
\resizebox{1.0\linewidth}{!}{
\setlength\tabcolsep{10pt}
    \begin{tabular}{lccccccc}
        \toprule
        \textbf{Model} & \multicolumn{1}{c}{Global} & \multicolumn{1}{c}{Entity} & \multicolumn{1}{c}{Attribute} & \multicolumn{1}{c}{Relation} & \multicolumn{1}{c}{Other} & \multicolumn{1}{c}{\textbf{Overall}} \\
        \midrule
        PixArt-$\alpha$~\citep{Chen2023PixArtFT} & 74.97 & 79.32 & 78.60 & 82.57 & 76.96 & 71.11 \\
        PlayGroundv2~\citep{playground-v2} & 83.61 &79.91 &82.67 &80.62 &81.22 & 74.54 \\
        PlayGroundv2.5~\citep{Li2024PlaygroundVT} & 83.06 & 82.59&81.20&84.08&83.50&75.47 \\
        Lumina-Next~\citep{Zhuo2024LuminaNextML} & 82.82	&88.65&86.44&80.53&81.82 & 74.63 \\
        DALLE-3~\citep{dalle3} & 90.97 & 89.61 & 88.39 & 90.58 & 89.83 & 83.50 \\
        SD3-medium~\citep{sd3} & 87.90 & 91.01 & 88.83 & 80.70 & 88.68 & 84.08 \\
        \midrule
        SDXL (baseline) & 82.44 & 81.87 & 81.17 & 80.54 & 79.77 & 73.38 \\ 
        \rowcolor{blue!5}
        SDXL-\ourmethod & 83.92\up{1.5} & 85.28\up{3.4} & 85.13\up{4.0} & 85.03\up{4.5} & 84.55\up{4.8} & 78.84\up{5.4} \\
        \bottomrule
    \end{tabular}
    }
\label{tab:dpg_result}
\end{table}

\begin{table}[t]
    \centering
    \caption{\textbf{Main Results on GenEval 2~\citep{Kamath2025GenEval2A}}.}
    \resizebox{0.7\linewidth}{!}{%
    \begin{tabular}{lcc}
        \toprule
        \textbf{Model} & \textbf{Soft-TIFA-AM}~$\uparrow$ & \textbf{Soft-TIFA-GM}~$\uparrow$ \\
        \midrule
        SDXL (baseline) & 50.1 & 9.1 \\
        \rowcolor{blue!5}
        SDXL-\ourmethod & 56.7\up{6.6} & 10.9\up{1.8} \\
        \bottomrule
    \end{tabular}
    }
    \label{tab:geneval2_result}
\end{table}

\begin{table}[t]
    \centering
    \caption{\textbf{Main Results on GenEval 2 (compositionality)~\citep{Kamath2025GenEval2A}}.}
    \resizebox{\linewidth}{!}{%
    \begin{tabular}{lcccccccc}
        \toprule
        \multirow{2}{*}{\textbf{Model}} & \multicolumn{8}{c}{\textbf{Atomicity~$\uparrow$}} \\
        \cmidrule(r){2-9}
        & 3 & 4 & 5 & 6 & 7 & 8 & 9 & 10 \\
        \midrule
        Flux & 48.0 & 28.0 & 16.0 & 26.0 & 4.0 & 0.0 & 0.0 & 0.0 \\
        SD3-Med. (baseline) & 52.0 & 30.0 & 16.0 & 8.0 & 0.0 & 4.0 & 2.0 & 0.0 \\
        \rowcolor{blue!5}
        SD3-Med.-\ourmethod & 52.2 & 40.8 & 18.3 & 27.6 & 14.2 & 9.3 & 7.0 & 3.3 \\
        \bottomrule
    \end{tabular}%
    }
    \label{tab:sd3_results}
\end{table}

\begin{table}[t]
    \centering
    \caption{\textbf{Visual aesthetic quality evaluation using HPSv2~\citep{Wu2023HumanPS}}.}
    \resizebox{\linewidth}{!}{%
    \begin{tabular}{l ccccc }
    \toprule
    \textbf{Model} & Concept-Art & Photo & Anime & Paintings & Average~$\uparrow$ \\
    \midrule
    SDXL & 30.42 & 27.97 & 31.71 & 30.76 & 30.22 \\
    \rowcolor{blue!5}
    SDXL-\ourmethod & 32.86\up{2.44} & 31.18\up{3.21} & 34.53\up{2.82} & 32.90\up{2.14} & 32.87~\up{2.65} \\
    \bottomrule
    \end{tabular}%
    }
    \label{tab:hpsv2_result}
\end{table}

\begin{table}[t]
\centering
\caption{\textbf{Ablation on key designs.} We report the overall scores over each benchmark.}
\resizebox{1.0\linewidth}{!}{
\setlength\tabcolsep{10pt}
    \begin{tabular}{lcccc}
        \toprule
        \textbf{Method} & \multicolumn{1}{c}{\textbf{T2I-CompBench}} & \multicolumn{1}{c}{\textbf{GenEval}} & \multicolumn{1}{c}{\textbf{DPG-Bench}} \\
        \cmidrule{1-1} \cmidrule{2-4}
        SDXL & 43.57 & 53.29 & 73.38 \\
        SDXL-SFT & 43.34 & 52.29 & 73.23 \\
        SDXL-ImageDPO & 45.58 & 53.00 & 75.70 \\
        SDXL-TextDPO & 13.48 & 4.71 & 23.98 \\
        SDXL-\ourmethod w/o region-level guidance & 53.10 & 60.71 & 77.53 \\
        \rowcolor{blue!5}
        SDXL-\ourmethod w/ region-level guidance & 54.37 & 62.14 & 78.84 \\
        \bottomrule
    \end{tabular}
    }
\label{tab:ablation_study}
\end{table}

\subsection{Experimental Setups}

\fakeparagraph{Implementation Details.}
We use Stable Diffusion XL (SDXL)~\citep{Podell2023SDXLIL} as our base model and fine-tune it with LoRA~\citep{lora} and set rank to 8. We train the model for 200 steps with an effective batch size equals to 2048. Learning rate is set to 2048 * 4e-8 with a constant schedule and 50 warm-up steps. All experiments are conducted on 4$\times$ H100 GPUs, with a total runtime of 13 hours.
For the region-level guidance, we set the weight to 1 for regions-of-interest and 0.5 for external regions to guide the model to focus on these regions. We do not use region-level guidance for data related to object numeracy or spatial relationships, as understanding these concepts requires a global focus.
For the training data, we use 53k samples in total, combining 42k from our \ourdataset dataset with 12k from VisMin~\citep{Awal2024VisMinVM} dataset.

\fakeparagraph{Evaluation Benchmarks.}
We evaluate the effectiveness of our method on four challenging benchmarks designed to assess compositional capabilities in text-to-image generation, \ie T2I-CompBench~\citep{huang2023t2icompbench}, GenEval~\citep{ghosh2023geneval}, DPG-Bench~\citep{dpgbench} and GenEval 2~\citep{Kamath2025GenEval2A}.

\subsection{Main Results}
\label{sec:main_exp}
\fakeparagraph{T2I-CompBench.} 
T2I-Compbench~\citep{huang2023t2icompbench} is a challenging benchmark that focuses on evaluating models in compositional generation, including object attributes and inter-object relationships.
As shown in Table~\ref{tab:t2i_compbench_result}, our method achieves significant improvements over the baseline SDXL model, especially in the attribute binding tasks (color, shape, texture). This demonstrates our method is effective in enhancing the model's ability to correctly associate attributes with their corresponding objects. 
Overall, our method achieves a substantial increase in the average score across all categories, highlighting its effectiveness for compositional text-to-image generation. 
Compared to other models designed for compositional generation, such as GLIGEN~\citep{li2023gligen}, LMD+~\citep{lmd}, and InstanceDiffusion~\citep{wang2024instancediffusion}, our model still demonstrates a clear advantage. It worth noting that these models require an additional layout condition for control, whereas \ourmethod achieves its strong performance using only the text prompts. 

\fakeparagraph{GenEval}.
We alse evaluate our \ourmethod on GenEval~\citep{ghosh2023geneval}, a benchmark designed to assess text-to-image models in complex instruction following.
As shown in \Cref{tab:geneval_result}, our \ourmethod achieves clear improvements over the SDXL baseline model across most of the sub-tasks. The overall score shows a notable increase (0.62 vs. 0.53), which demonstrates our method's effectiveness in enhancing the base model to follow complex text prompts.
Furthermore, our method even surpasses state-of-the-art models such as DALL-E 3~\citep{dalle3} and FLUX.1-dev~\citep{flux} in several sub-tasks, including ``single object'' and ``colors''. This is particularly notable given our model is significantly smaller size and is trained on substantially less data. 

\fakeparagraph{DPG-Bench}
We also evaluate our method on DPG-Bench~\citep{dpgbench}, a comprehensive benchmark for assessing the intricate semantic alignment capabilities of text-to-image models.
As illustrated in \Cref{tab:dpg_result}, our \ourmethod-SDXL achieves competitive results on the benchmark. Specifically, our model obtains comparable scores across all categories, including Global (83.92), Entity (85.28), Attribute (85.13), Relation (85.03), and Other (84.55), with a strong overall score of 78.84. Compared to the SDXL baseline (73.38 overall), our method demonstrates clear improvements, particularly in the Entity, Attribute, and Relation categories. These results validate the effectiveness as well as the robustness of our approach for compositional text-to-image generation.

\fakeparagraph{GenEval 2.}
We further evaluate our method on GenEval 2~\citep{dpgbench}, a more challenging benchmark well-suited for modern models. As shown in~\Cref{tab:geneval2_result}, compared to the baseline, \ourmethod exhibits significant improvements in both atomic-level (6.6\%) and prompt-level (1.8\%). This demonstrate that our \ourmethod is robust to benchmark drift.

\fakeparagraph{Extending \ourmethod to Modern MMDiT.}
We conduct experiments on the prevailing MMDiT architecture. As shown in~\Cref{tab:sd3_results}, \ourmethod brings significant improvements to SD3-Medium, particularly as compositional complexity increases. Notably, with the assistance of \ourmethod, SD3-Medium even outperforms Flux. This validates that \ourmethod is model-agnostic and can generalize well to current SOTA models.

\fakeparagraph{Visual Aesthetic Quality Evaluation.}
We use HPSv2 as aesthetic assessment metric and evaluate on DrawBench. As shown in~\Cref{tab:hpsv2_result}, \ourmethod achieves a 2.65\% improvement in visual quality. This indicates that \ourmethod enhances visual quality while improving compositionality.

\subsection{Ablation Studies}

\label{sec:ablate_exp}
We conduct extensive ablation studies to evaluate the key designs of \ourmethod. We use SDXL as our baseline model, and explore several fine-tuning configurations:
\begin{itemize}[leftmargin=1em]
    \item \textbf{SFT}: Supervised fine-tuning without any kind of preference optimization.
    \item \textbf{ImageDPO}: Applying DPO using only image preferences (positive and negative images).
    \item \textbf{TextDPO}: Applying DPO using only text preferences (positive and negative texts).
    \item \textbf{\ourmethod} (w/o region-level guidance): Applying bimodal DPO, using both positive and negative images and texts.
    \item \textbf{\ourmethod} (w/ region-level guidance): Bimodal DPO with region-level guidance based on bounding box annotations.
\end{itemize}

\fakeparagraph{Effectiveness of Bimodal Preference Optimizing.}
As shown in ~\Cref{sec:ablate_exp}, directly performing supervised fine-tuning on the composition-aware dataset fails to guide the model to focus on attribute binding and object relationships, demonstrating the necessity of preference optimization. In contrast, ImageDPO achieves a certain degree of performance improvement. This highlights the importance of guiding the model to focus on fine-grained compositional attributes through the comparison between positive and negative examples via direct preference optimization. However, solely perform text comparison leads to significant performance drop, as it lacks visual guidance for \textbf{generation} and fails to provide effective supervision on visual details. TextDPO lacks visual guidance for \textbf{generation} and fails to provide effective supervision on visual details, result in visual quality degradation. In contrast, simultaneously optimizing preferences from both images and text more effectively promotes the model's cross-modal alignment, leading to a highly significant performance improvement.

\fakeparagraph{Effectiveness of Region-level Guidance.}
From the last two lines of ~\Cref{tab:ablation_study}, it can be observed that the introduction of region-level guidance on top of \ourmethod leads to further improvements (1.2\% on T2I-CompBench and 1.4\% on GenEval). This indicates that explicitly guiding the model to focus on regions in the image that are relevant to the text description can effectively enhance the models to achieve fine-grained cross-modal alignment.

\section{Conclusion}
In this work, we present \ourmethod, a novel method that introduces DPO to compositional text-to-image generation, extends it to a bimodal version and further enhances it with region-level guidance. Trained on our created composition-aware preference dataset \ourdataset, \ourmethod significantly improves the compositional capabilities of text-to-image diffusion models, as demonstrated by extensive experiments on four standard benchmarks: T2I-CompBench, GenEval, DPG-Bench and GenEval 2. For future work, we plan to extend our method to more kinds of text-to-image models like autoregressive models.
\newpage
\fakeparagraph{Acknowledgments.}
This work was supported by by National Natural Science Foundation of China (No. 62521004) and the Science and Technology Commission of Shanghai Municipality (No. 25511106100).
\section*{Appendix}
In this supplementary, we provide additional details and results as follows:
\begin{itemize}[leftmargin=0.5em]
    \item In \Cref{sec:ablation_study_details}, we provide the full results of the ablation study on SDXL based models.
    \item In \Cref{sec:data_construction_details}, we provide more details about our data construction process, including the composition of collected captions from various sources and the prompts used in various stages.
    \item In \Cref{sec:more_visualization_results}, we provide more visualization results of our \ourdataset dataset and our \ourmethod method.
\end{itemize}

\section{Ablation Study Details.}
\label{sec:ablation_study_details}
The full results of the ablation study on SDXL based models are shown in \Cref{tab:t2i_compbench_abl}, \Cref{tab:geneval_abl} and \Cref{tab:dpg_abl}.

\begin{table}[h]
\centering
\caption{\textbf{Ablation Study on T2I-CompBench~\citep{huang2023t2icompbench}}.}
\resizebox{\linewidth}{!}{
\begin{tabular}{l ccc cc c}
\toprule
\multirow{2}{*}{\textbf{Model}} & 
\multicolumn{3}{c}{\textbf{Attribute Binding}} &
\multicolumn{2}{c}{\textbf{Object Relationship}} &
\multirow{2}{*}{\textbf{Numeracy}} \\
\cmidrule(lr){2-4} \cmidrule(lr){5-6}
& Color & Shape & Texture & Spatial & Non-Spatial & \\
\midrule
SDXL & 58.90 & 46.90 & 53.13 & 21.23 & 31.20 & 50.08\\
SDXL-SFT & 58.67 & 46.65 & 52.21 &  21.13 & 31.28 & 50.08\\
SDXL-ImageDPO & 67.39 & 53.12 & 59.42 & 23.4  & 30.82 & 39.34\\
SDXL-TextDPO & 23.32  & 16.22 & 14.62 & 0.26  & 20.3 & 6.13\\
SDXL-\ourmethod w/o region-level guidance & 77.04  & 57.43 & 68.89 & 23.19  & 32.19 & 59.83 \\
SDXL-\ourmethod w/ region-level guidance & 79.35 & 60.47 & 71.36 & 23.41 & 32.29  & 59.33\\
\bottomrule
\end{tabular}
}
\label{tab:t2i_compbench_abl}
\end{table}

\begin{table}[h]
    \centering
    \caption{\textbf{Ablation Study on GenEval~\cite{ghosh2023geneval}}.}
    \resizebox{\linewidth}{!}{
    \begin{tabular}{lcccccccc}
        \toprule
        \multicolumn{1}{c}{\textbf{Model}} & \multicolumn{1}{c}{Single Obj.} & Two Obj. & Counting & Colors & Position & Color Attri. & \textbf{Overall} \\
        \midrule
        SDXL &   0.95 & 0.68 & 0.42 & 0.85 & 0.11 & 0.19 & 0.53  \\
        SDXL-SFT & 0.95 & 0.68 & 0.37 & 0.85 & 0.09 & 0.20 & 0.52  \\
        SDXL-ImageDPO  & 0.99 & 0.78 & 0.15 & 0.89 & 0.14 & 0.23 & 0.53  \\
        SDXL-TextDPO  & 0.13 & 0.01 & 0.01 & 0.11 & 0.01 & 0.02 & 0.04  \\
        SDXL-\ourmethod w/o region-level guidance & 1.00 & 0.83 & 0.52 & 0.90 & 0.16 & 0.23 & 0.61  \\
        SDXL-\ourmethod w/ region-level guidance & 1.00  & 0.87  & 0.56  & 0.90  & 0.17  & 0.23  & 0.62 & \\
        \bottomrule
    \end{tabular}
    }
    \label{tab:geneval_abl}
\end{table}

\begin{table}[h]
\centering
\caption{\textbf{Ablation Study on DPG-Bench~\cite{dpgbench}}.}
\resizebox{1.0\linewidth}{!}{
\setlength\tabcolsep{10pt}
    \begin{tabular}{lccccccc}
        \toprule
        \textbf{Model} & \multicolumn{1}{c}{Global} & \multicolumn{1}{c}{Entity} & \multicolumn{1}{c}{Attribute} & \multicolumn{1}{c}{Relation} & \multicolumn{1}{c}{Other} & \multicolumn{1}{c}{Overall} \\
        \midrule
        SDXL & 82.44 & 81.87 & 81.17 & 80.54 & 79.77 & 73.38 \\ 
        SDXL-SFT & 82.68 & 81.94 & 79.52 & 81.02 & 81.03 & 73.23 \\
        SDXL-ImageDPO & 79.13 & 83.19 & 82.76 & 83.39 & 82.49 & 75.70 \\
        SDXL-TextDPO & 40.07 & 40.44 & 43.14 & 43.92 & 44.82 & 23.98 \\
        SDXL-\ourmethod w/o region-level guidance & 85.46 & 84.22 & 84.45 & 85.28 & 84.18 & 77.53 \\
        SDXL-\ourmethod w/ region-level guidance & 83.92 & 85.28 & 85.13 & 85.03 & 84.55 & 78.84 \\
        \bottomrule
    \end{tabular}
    }
\label{tab:dpg_abl}
\end{table}

\section{Data Construction Details}
\label{sec:data_construction_details}

\fakeparagraph{Caption Collection.}
~\Cref{tab:caption-dataset} shows the number of captions collected from each source.

\begin{table}[h]
    \centering
    \begin{tabular}{l c}
        \toprule
        \textbf{Dataset} & \textbf{Number of Captions} \\
        \midrule
        CONPAIR~\citep{contrafusion} & 13,432 \\
        ReasonGen-R1~\citep{Zhang2025ReasonGenR1CF} & 23,470 \\
        T2I-R1~\citep{Jiang2025T2IR1RI} & 7,223 \\
        T2I-CompBench Training Set~\citep{huang2023t2icompbench} & 5,600 \\
        \midrule
        \textbf{Total} & \textbf{49,725} \\
        \bottomrule
    \end{tabular}
    \caption{Number of composition-related captions collected from various sources.}
    \label{tab:caption-dataset}
\end{table}

\fakeparagraph{LLM Prompt for Dimension Parsing.}
Our dimension parsing prompt is shown in Listing \ref{dimension_parsing_prompt}. We use DeepSeek-V3 ~\citep{DeepSeekAI2024DeepSeekV3TR} as our LLM to parse the captions. 

\lstset{
    backgroundcolor=\color{gray!10},
    basicstyle=\ttfamily\small,
    frame=single,
    breaklines=true,
    postbreak=\mbox{\textcolor{red}{$\hookrightarrow$}\space}
}

\begin{lstlisting}[caption={prompt for dimension parsing}, label=dimension_parsing_prompt]
# Task Description
Given a sentence, analyze its content and determine which of the following dimensions it primarily describes:
- color: describes colors (e.g., "red", "yellow", "dark blue")
- shape: describes geometric forms or outlines (e.g., "round", "triangular", "curved")
- texture: describes textures (e.g., "smooth", "rough") or materials (e.g., "a plastic chair", "a glass window")
- spatial: describes spatial relationships or positions (e.g., "on the table", "next to", "inside", "beneath")
- non-spatial: describes actions/events without spatial focus (e.g., "chasing", "biting")
- numeracy: describes quantities or numbers (e.g., "three apples", "four", "two")
- others: when none of the above categories apply

# Priority Rules
If multiple dimensions are present, select according to this priority:
1. spatial and non-spatial have highest priority (equal)
2. numeracy comes next
3. color, shape and texture have equal priority (lower than above)
4. others is always lowest priority

# Output Format
Provide your analysis in exact JSON format as shown below. Only include the JSON object in your response.

{{
    "dimension": "selected_dimension"
}}

# Examples
Input: "The cube is on the shelf"
Output: {{ "dimension": "spatial" }}

Input: "Five rough textured stones"
Output: {{ "dimension": "numeracy" }}

Input: "The soft yellow pillow"
Output: {{ "dimension": "color" }}

# Input
The input sentence is: {positive_caption}

# Output
For this sentence, the dimension is:
\end{lstlisting}

\fakeparagraph{LLM Prompt for Object List Parsing.}
We use the prompt shown in Listing \ref{object_list_parsing_prompt} to parse the object list from captions. We use DeepSeek-R1 ~\citep{DeepSeekAI2025DeepSeekR1IR} as our LLM to parse the captions.

\begin{lstlisting}[caption={prompt for object list parsing}, label=object_list_parsing_prompt]
You are an expert in parsing textual sentences. Given a text that describing an image, you task is to identify and extract the main entities in the image.

# Requirements
- You should only put the main entities that are visually visible in the image.
- Make sure the entities you identify are concrete objects, not abstract concepts; objects like 'living room' or 'wind' should not be identified.
- Make sure these entity objects can be detected by an object detector.
- Only output the entities themselves, without their adjectives or descriptions; for example, output 'dog' instead of 'white dog'.

# Output format
Orgainize the identified main objects in the scene into a json dict like this:
{{
    "object_list": ["object 1", "object 2", ...]
}}
# Input
For the sentence: {caption}, please identify the main visible objects.
\end{lstlisting}

\fakeparagraph{Image Describing Details.}
Before we prompt the VLM to do the describing tasks, we restrict the image to follow the following rules:
1) with dimension "color", "shape", or "texture", the image should contain one or two objects
2) with dimension "spatial" or "non-spatial", the image should contain exactly two objects.
3) no repeated classes in the image; each object must belong to a unique class.
We use specific prompts for different dimensions. Examples of ``color'' and ``spatial'' dimension are shown in Listing \ref{color_prompt} and Listing \ref{spatial_prompt}. The prompts for ``shape'', ``texture'', and ``non-spatial'' dimensions are similar to the ``color'' and ``spatial'' ones, respectively.
We use Qwen2.5-VL-72B-Instruct~\citep{Bai2025Qwen25VLTR} as our VLM to describe the images.

\begin{lstlisting}[caption={prompt for VLM describing, with dimension ``color''}, label=color_prompt]
# Task explanation
Given an image with clearly marked regions-of-interest (each region is indicated by a numerical ID and contour lines), please:
1. Identify all visible regions-of-interest by their numerical IDs
2. For each region, determine the predominant color of the object contained within it
3. Describe colors using standard web color names (e.g., "red", "forestgreen", "royalblue")
4. Handle uncertainty cases appropriately

# Output Requirements:
- Strict JSON format
- For unclear cases: use "unknown" as color value
- Sort results by region ID in ascending order

Output Example:
{
  "color_predictions": [
    {
      "region_id": 1,
      "color": "red"
    },
    {
      "region_id": 2,
      "color": "unknown"
    }
  ]
}

# Special Instructions:
- Ignore background colors outside marked regions
- Focus on the dominant colors
- IDs and contour lines are only for reference. DO NOT use them for color analysis
\end{lstlisting}

\begin{lstlisting}[caption={prompt for VLM describing, with dimension ``spatial''}, label=spatial_prompt]
# Task explanation
Given an image with two clearly marked regions-of-interest (each region is indicated by a numerical ID and contour lines), please:
1. Identify the two regions-of-interest by their numerical IDs
2. Determine the precise spatial relationship between the two objects contained within the two regions-of-interest, where:
   - The reference object should be the visually more salient/dominant object (typically larger, more central, or more prominent in the scene)
   - The target object's position is described relative to the reference object
   - Use specific spatial descriptors (e.g., "on the right of", "above", "behind")
3. Handle uncertainty cases appropriately when spatial relationships cannot be clearly determined

# Output Requirements:
- Strict JSON format
- For unclear spatial relationship: use "unknown"
- Always describe the target object's position relative to the reference object

Output Example:
{
  "reference_object_id": 1,
  "target_object_id": 2,
  "spatial_prediction": "in front of", 
  "notes": "object 2 is in front of object 1"
}

For unknown cases:
{
  "spatial_prediction": "unknown"
}

# Special Instructions:
- Do not use unclear descriptions like "next to", "beside", "near", "close to", etc
- IDs and contour lines are only for reference. DO NOT use them for spatial relationship analysis
- If neither object is clearly more salient, default to using the lower ID as reference
\end{lstlisting}

\fakeparagraph{VLM prompts for Region Information Differing.}
We use specific prompt for each dimension to generate distinct region information. Examples of ``color'' and ``spatial'' dimension are shown in Listing \ref{color_prompt_diff} and Listing \ref{spatial_prompt_diff}. The prompts for ``shape'', ``texture'', and ``non-spatial'' dimensions are similar to the ``color'' and ``spatial'' ones, respectively.
We use Qwen2.5-VL-72B-Instruct~\citep{Bai2025Qwen25VLTR} as our VLM to generate the distinct region information.

\begin{lstlisting}[caption={prompt for VLM differentiation, with dimension ``color''}, label=color_prompt_diff]
# Task Explanation
Here is an image with outlined regions (each region is indicated by a numerical ID and contour lines).
And here are the list of regions with their dominant colors for reference (format: {{"region_id": N, "color": "color_name"}}):
{obj}

Now please propose a visually distinct color for each region that significantly differs from ALL provided dominant colors in the image.

# Requirements:
1. For each region, suggest ONE color that contrasts distinctly with ALL dominant colors in the image
2. Consider human perceptual difference (avoid suggesting similar hues/brightness)
3. Prefer standard color names (e.g., "red", "green")
4. Never suggest the same as any input dominant color
5. When multiple options exist, choose the highest-contrast alternative

# Output Format (strict JSON):
{{
  "output": [
    {{"region_id": N, "different_color": "color_name"}},
    ...(other regions)
  ]
}}

# Examples:
Input Colors: [{{"region_id": 1, "dominant_color": "red"}}, {{"region_id": 2, "dominant_color": "blue"}}]
Output: {{
  "output": [
    {{"region_id": 1, "different_color": "yellow"}},
    {{"region_id": 2, "different_color": "black"}}
  ]
}}

Input Colors: [{{"region_id": 1, "dominant_color": "green"}}, {{"region_id": 2, "dominant_color": "yellow"}}]
Output: {{
  "output": [
    {{"region_id": 1, "different_color": "magenta"}},
    {{"region_id": 2, "different_color": "navy_blue"}}
  ]
}}
\end{lstlisting}

\begin{lstlisting}[caption={prompt for VLM differentiation, with dimension ``spatial''}, label=spatial_prompt_diff]
# Task Explanation
Here is an image with TWO outlined regions (each region is indicated by a numerical ID and contour lines).
And here is the spatial relationship between the two objects contained in the two regions:
object {object_id_1} is {spatial_relation} object {object_id_2}

Now please propose a geometrically distinct spatial relationship that significantly differs from the given relationship.

# Requirements:
1. Suggest ONE primary spatial relationship that contrasts maximally with the input relationship
2. Consider these transformation axes for differentiation:
   a) Vertical inversion (above/below → swap)
   b) Horizontal inversion (left/right → swap) 
   c) Dimensional shift (adjacent → separated)
   d) Topological change (inside → outside)
3. Use standard spatial terms from this vocabulary:
   [above, below, on the left of, on the right of, in front of, behind, ...]
4. The new relationship must be:
   a) Physically plausible for the objects' shapes/sizes
   b) Perceptually distinct from original
   c) Expressed as "object X [RELATION] object Y"
5. Include brief reasoning in "notes"

# Output Format (strict JSON):
{{
  "output": {{
    "different_spatial_relation": "relation_term",
    "notes": "object [object_id_X] [RELATION] object [object_id_Y]"
  }}
}}

# Examples:
Input: "object A is above object B"
Output: {{
  "output": {{
    "different_spatial_relation": "below",
    "notes": "object A is below object B (vertical inversion)"
  }}
}}

Input: "object X is inside object Y"
Output: {{
  "output": {{
    "different_spatial_relation": "outside",
    "notes": "object X is outside object Y (topological complement)"
  }}
}}

Input: "object 1 is adjacent to object 2"
Output: {{
  "output": {{
    "different_spatial_relation": "separated",
    "notes": "object 1 is separated from object 2 (proximity reversal)"
  }}
}}
\end{lstlisting}

\fakeparagraph{VLM prompt for VQA-based filtering.}
We use the prompt shown in Listing \ref{vqa_prompt} to filter out low-quality samples.
We use Qwen2.5-VL-72B-Instruct~\citep{Bai2025Qwen25VLTR} as our VLM to perform the filtering.

\begin{lstlisting}[caption={prompt for VLM VQA-based filtering}, label=vqa_prompt]
You are given an image with several regions of interest (ROIs). Each ROI is highlighted in the image with contour lines and labeled with a unique numerical ID.

You are also given a list of questions. Each question refers to one or more ROIs. Here are the questions:
{questions}

Your task:

1. For each question, evaluate whether the statement is correct with respect to the corresponding region(s).
2. Provide a confidence score between 0 and 1 (`answer`) indicating how strongly you agree with the statement (1 = completely true, 0 = completely false).
3. Provide a short explanation (`reason`) describing why you assigned this score.

The output format must strictly follow this JSON structure:

```json
[
  {{
    "question_id": <int>,
    "answer": <float between 0 and 1>,
    "reason": "<string explanation>"
  }},
  ...
]
```

**Example:**
Input image: contains region 1 (a yellow lemon) and region 2 (a red apple).
Questions:

```json
[
  {{"question_id": 0, "question": "Does region 1 mark a yellow lemon?"}},
  {{"question_id": 1, "question": "Does region 2 mark a blue apple?"}}
]
```

Expected output:

```json
[
  {{"question_id": 0, "answer": 0.99, "reason": "Region 1 does mark a yellow lemon."}},
  {{"question_id": 1, "answer": 0.01, "reason": "The apple in region 2 is actually red."}}
]
```
\end{lstlisting}

\section{More Visualization Results.}
\label{sec:more_visualization_results}
\begin{figure}[h]
    \centering
    \includegraphics[width=0.8\textwidth]{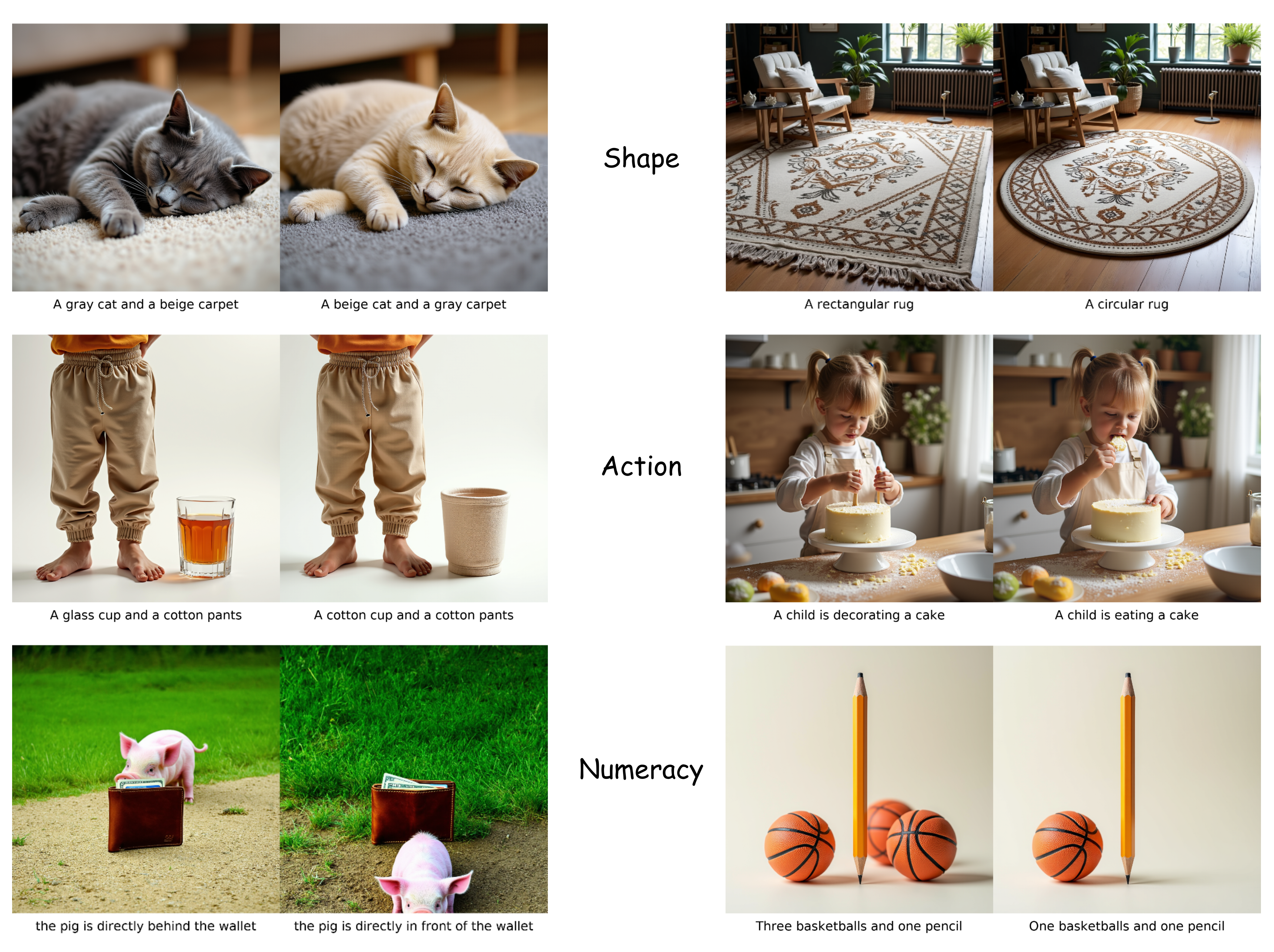}
    \caption{\textbf{Samples of each dimension in our \ourdataset dataset.} For each group, the left image is generated from the original caption, and the right image is generated from the edited caption.}
    \label{fig:data_samples}
\end{figure}

\begin{figure}[h]
    \centering
    \includegraphics[width=0.8\textwidth]{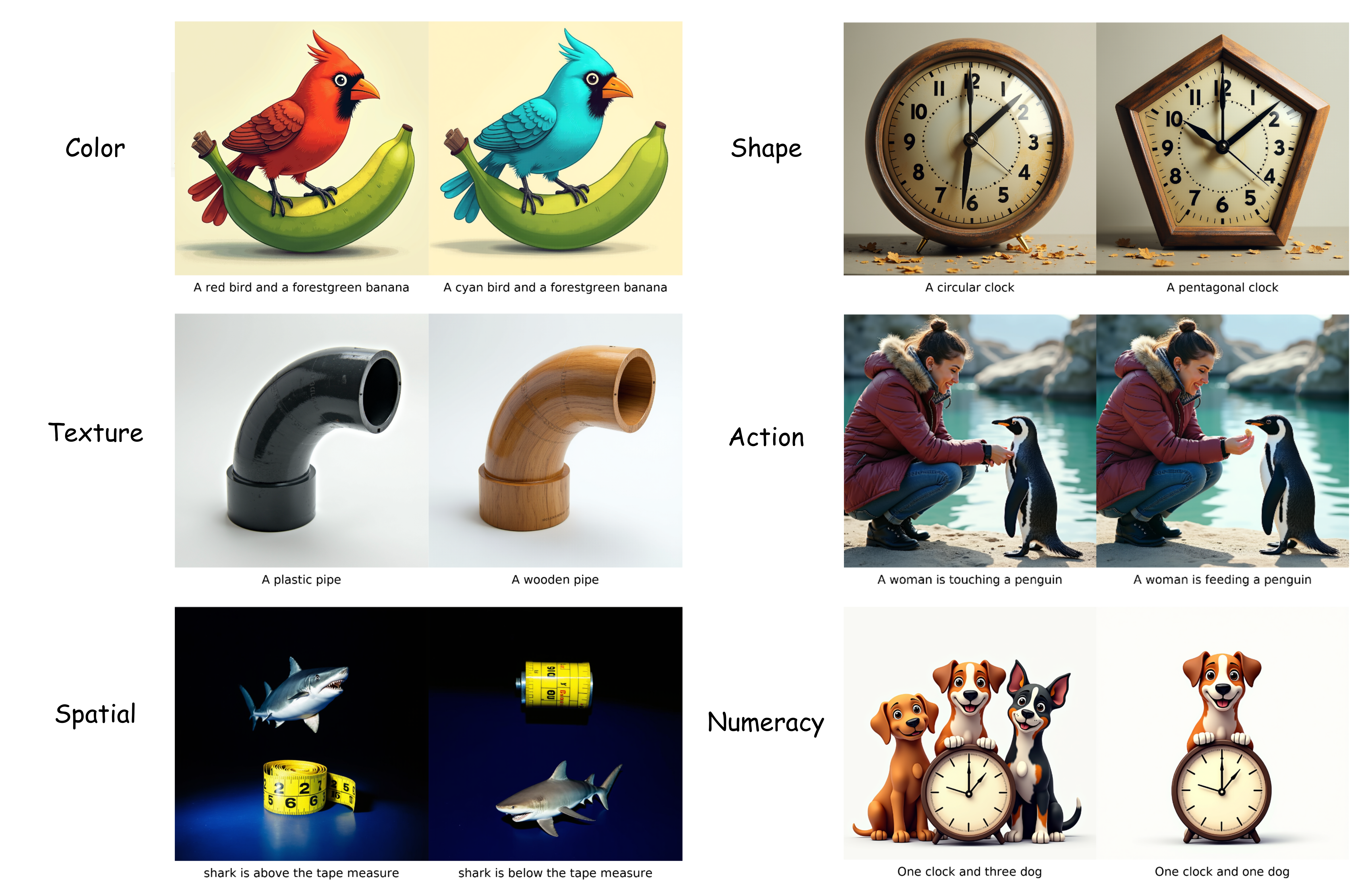}
    \caption{\textbf{Samples of each dimension in our \ourdataset dataset}. For each group, the left image is generated from the original caption, and the right image is generated from the edited caption.}
    \label{fig:datasample2}
\end{figure}

\begin{figure}[h]
    \centering
    \includegraphics[width=0.8\textwidth]{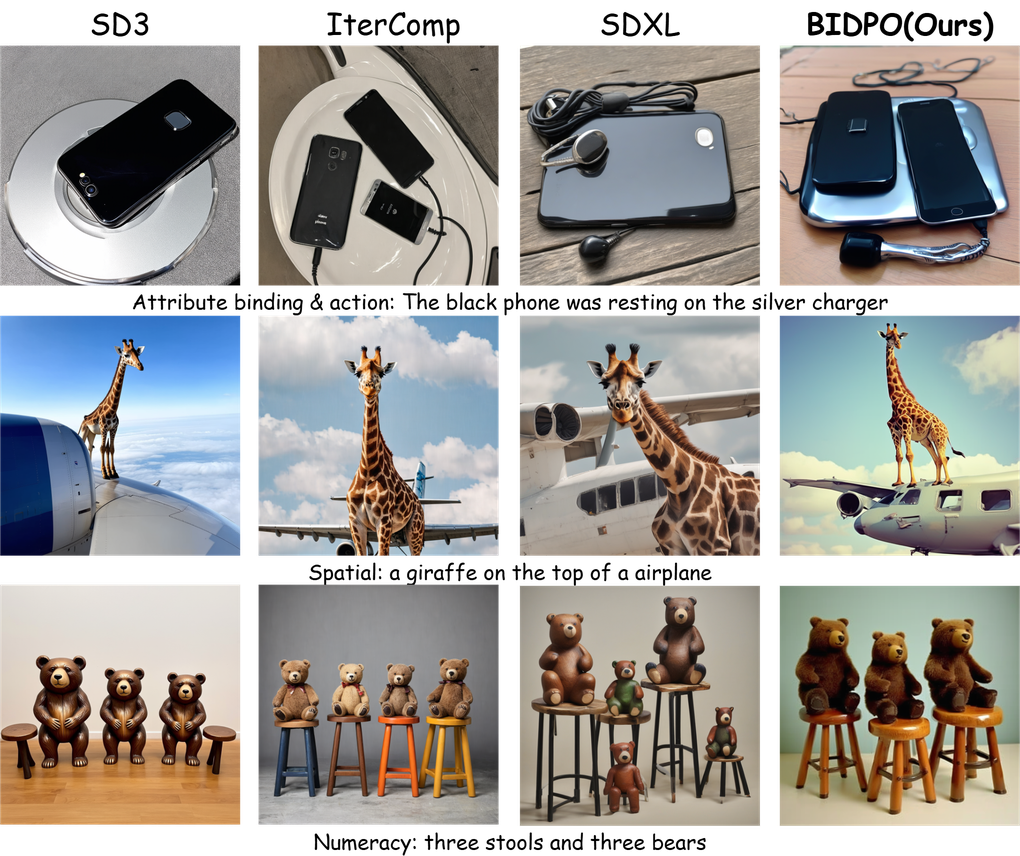}
    \caption{\textbf{Visualization of text-to-image generation results}. From left to right are Stable Diffusion 3 \citep{sd3}, IterComp \citep{zhang2024itercomp}, Stable Diffusion XL \citep{Podell2023SDXLIL}, and Stable Diffusion XL finetuned with our proposed \ourmethod.}
    \label{fig:sdxlteaser}
\end{figure}

\begin{figure}[h]
    \centering
    \includegraphics[width=0.8\textwidth]{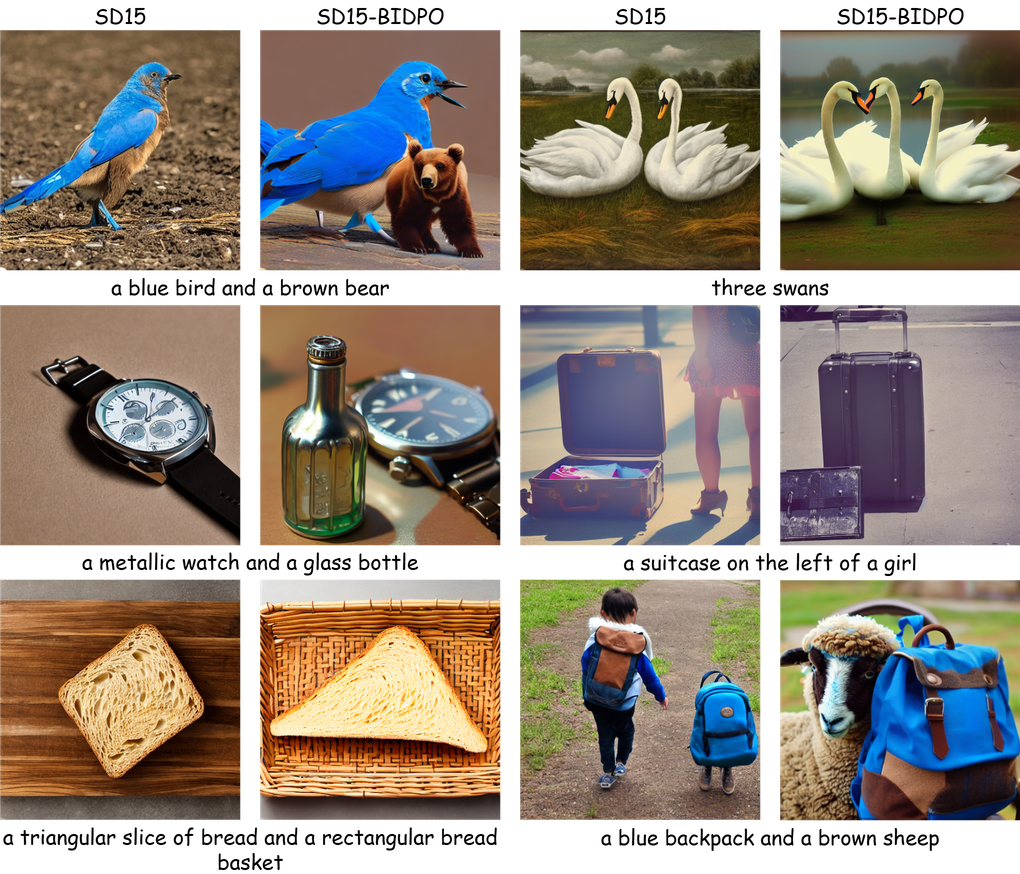}
    \caption{\textbf{Visualization of text-to-image generation results of Stable Diffusion 1.5}. We compare Stable Diffusion 1.5 finetuned with our proposed \ourmethod with the original Stable Diffusion 1.5.}
    \label{fig:sd15teaser}
\end{figure}

\begin{figure}[h]
    \centering
    \includegraphics[width=0.8\textwidth]{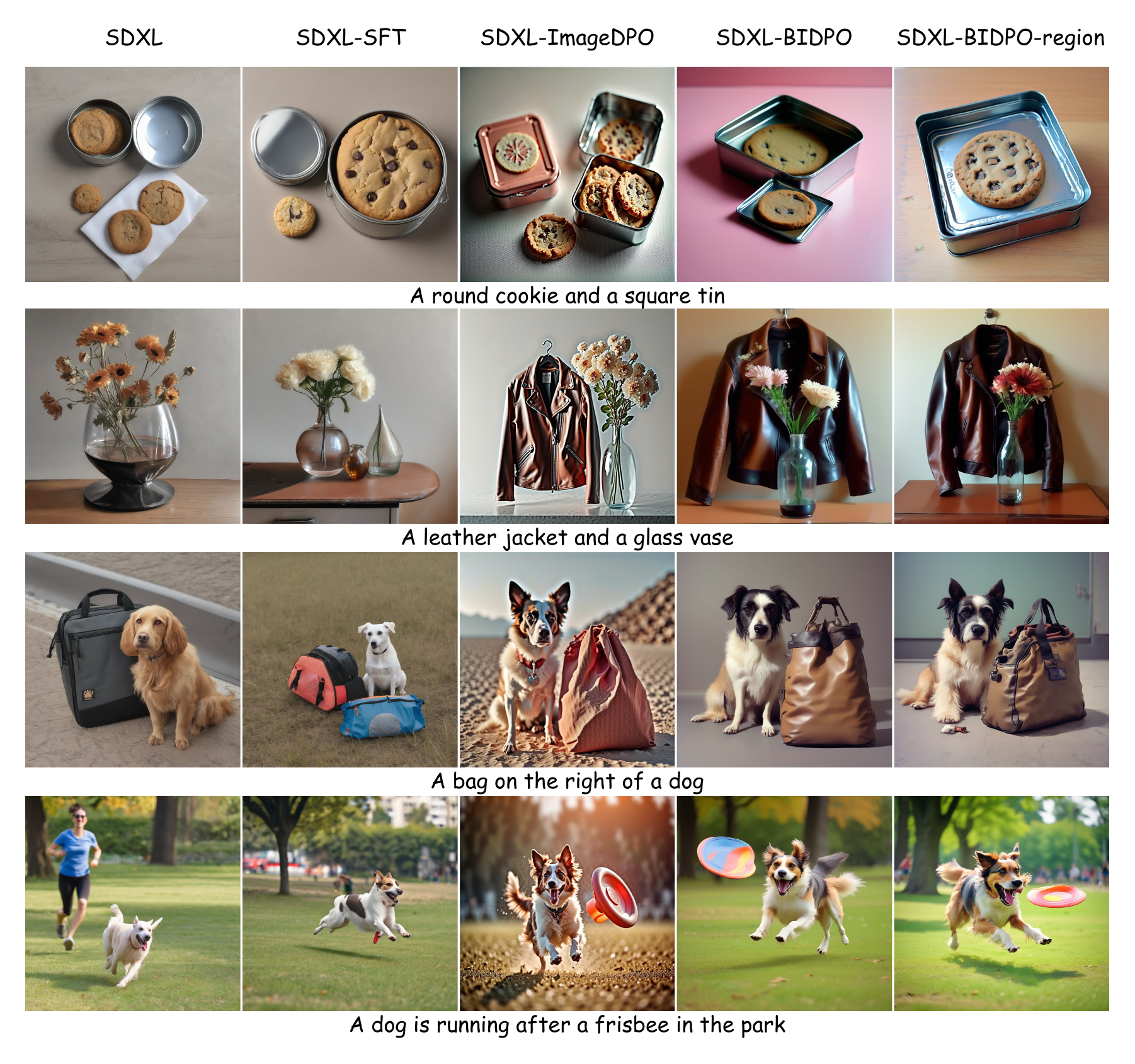}
    \caption{\textbf{Visualization of text-to-image generation results of Stable Diffusion XL}.}
    \label{fig:sdxl_visual}
\end{figure}

We provide more visualization results of our \ourdataset dataset and our \ourmethod method in \Cref{fig:data_samples}, \Cref{fig:datasample2}, \Cref{fig:sdxlteaser}, \Cref{fig:sd15teaser} and \Cref{fig:sdxl_visual}.

\clearpage

\bibliographystyle{plainnat}
\bibliography{main}

\end{document}